\documentclass[journal]{IEEEtran}
\usepackage[square,sort,comma,numbers]{natbib}

\ifCLASSINFOpdf
\else
\fi
\usepackage{amsmath}
\usepackage{amssymb}
  {

  }

\usepackage{hhline}
\usepackage[hyphens]{url}  
\usepackage{subcaption}
\usepackage{tcolorbox}
\usepackage{color}
\usepackage{colortbl}
\usepackage[makeroom]{cancel}
\definecolor{myred}{rgb}{0.8,0,0}
\definecolor{mygreen}{rgb}{0,0.6,0}
\definecolor{myblue}{rgb}{0,0,0.7}

\definecolor{DarkGray}{gray}{0.9}
\definecolor{MediumGray}{gray}{0.75}
\definecolor{LightGray}{gray}{0.5}

\usepackage{algorithm}
\usepackage[noend]{algpseudocode}

\def\bbbr{{\rm I\!R}}

\newcommand{\R}{{\bbbr}{}}

\usepackage{array}
\newcolumntype{P}[1]{>{\centering\arraybackslash}p{#1}}
\newcolumntype{M}[1]{>{\centering\arraybackslash}m{#1}}
\begin{document}
%
\title{Leveraging Sequentiality in Reinforcement Learning from a Single Demonstration}
%
%
%

\author{Alexandre~Chenu$^{1}$,
        Olivier~Serris$^{1}$,
        Olivier~Sigaud$^{1}$,
        and~Nicolas~Perrin-Gilbert$^{1}$
\thanks{$^{1}$Sorbonne Université, CNRS, Institut des Systèmes Intelligents et de Robotique, ISIR F-75005 Paris, France \textit{chenu@isir.upmc.fr}}
}

\maketitle

\begin{abstract}

Deep Reinforcement Learning has been successfully applied to learn robotic control.
However, the corresponding algorithms struggle when applied to problems where the agent is only rewarded after achieving a complex task. 
In this context, using demonstrations can significantly speed up the learning process, but demonstrations can be costly to acquire. 
In this paper, we propose to leverage a sequential bias to learn control policies for complex robotic tasks using a single demonstration. To do so, our method learns a goal-conditioned policy to control a system between successive low-dimensional goals.
This sequential goal reaching approach raises a problem of compatibility between successive goals: we need to ensure that the state resulting from reaching a goal is compatible with the achievement of the following goals.
To tackle this problem, we present a new algorithm called DCIL-II.
We show that DCIL-II can solve with unprecedented sample efficiency some challenging simulated tasks such as humanoid locomotion and stand-up as well as fast running with a simulated Cassie robot. Our method leveraging sequentiality is a step towards the resolution of complex robotic tasks under minimal specification effort, a key feature for the next generation of autonomous robots. 


\end{abstract}


%
\IEEEpeerreviewmaketitle

\section{Introduction}
%
%
%
%







Specifying the behavior of robots to achieve complex tasks is often difficult. Given the recent successes of Deep Reinforcement Learning (DRL) methods in robotics \cite{akkaya2019solving}, it is tempting to call upon these methods to circumvent the heavy burden of complex behavior specification. The approach would consist in just defining a success criterion as a reward function which would be positive when the task is achieved and null otherwise, and let the DRL algorithm discover the  so-defined successful behavior. Unfortunately, it happens that DRL algorithms struggle to solve precisely these tasks where the target behavior is complex and the reward signal corresponding to success is sparse. For instance, an off-the-shelf DRL algorithm cannot learn a control policy for verticalizing an underactuated Humanoid robot if the reward signal is only received when the robot is standing.

\begin{figure}[ht!]
     \centering
     \includegraphics[width = 0.9\hsize]{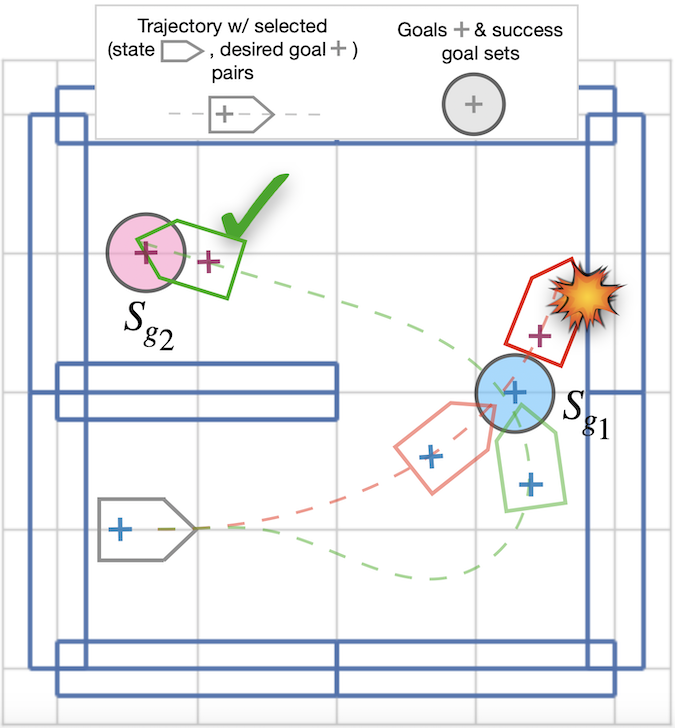}
    \caption{Illustration of the sequential goal-reaching  problem when using low-dimensional projections of states as goals. In this toy 2D maze, the agent corresponds to a Dubins car \cite{dubins1957curves} with $(x,y,\theta)$ states. Goals $g_{i}$ correspond to desired $(x,y)$ positions but do not condition the orientation $\theta$ of the car. In the red trajectory, the car reaches goal $g_{1}$ with an orientation that is not compatible with the achievement of goal $g_{2}$. On the contrary, in the green trajectory, the car trajectory is adapted to reach the first goal with a valid orientation. The GCRL framework we present in this paper is designed to learn policies that ensure targeting valid states when achieving a sequence of low-dimensional goals extracted from a demonstration.}
    \label{fig:sequential_goal_reaching_problem} 
\end{figure}

In such a context, leveraging expert demonstrations to bootstrap the learning process is a fruitful approach, as it efficiently accelerates policy optimization \cite{ho2016generative,kostrikov2018discriminator}. In Imitation Learning (IL), the demonstrations are used to match the agent's behavior to the expert's one.
However, many demonstrations containing both states and actions are generally required to successfully perform IL \cite{pomerleau1991efficient, russell1998learning}. 
Obtaining such demonstrations can be prohibitively difficult when dealing with complex systems in real world environments. Moreover, actions are not always available. For instance, when demonstrations come from motion capture \cite{merel2018hierarchical, peng2018deepmimic} or human guidance \cite{johns2021coarse}, only state-based trajectories are available.

In this paper, we focus on the problem of learning a control policy to solve complex robotics tasks with a very sparse reward (or in the absence of any reward) using a single demonstration. We tackle this problem by leveraging a sequential inductive bias in the way we model the task. 
First, we convert the demonstration into a succession of intermediate low-dimensional goals. 
Then, we use goal-conditioned reinforcement learning (GCRL) to learn a policy able to control the robot to sequentially reach the intermediate goals. 
For instance, in the Humanoid verticalization scenario above, a sequence of goals may correspond to a succession of Cartesian positions of the torso of the Humanoid from the ground to a desired height. By achieving these goals sequentially, the robot can stand up.

However, when dealing with a high-dimensional state space, reaching a low-dimensional goal does not fully condition the current state of the agent. This can raise issues if some intermediate goal reaching states are incompatible with reaching the next goal. 
For instance, consider a simple 2D kinematic car with non-holonomic constraints like the Dubins car \cite{dubins1957curves} navigating a maze, as illustrated in \figurename~\ref{fig:sequential_goal_reaching_problem}. In this environment, reaching a desired $(x,y)$ position does not condition the arrival orientation $\theta$ of the car. This partial conditioning of the success state may be problematic when we want to reach successive $(x,y)$ positions. Indeed, if the car reaches a goal close to a wall and is oriented towards it, the car may not escape from hitting the wall before reaching the next goal.

Thus, when considering a sequence of goals corresponding to low-dimensional projections of high-dimensional states, it is essential to ensure that the agent achieves each goal by reaching states compatible with the achievement of the next goal, that we call {\em valid success states}. This is the key property ensured by the DCIL-II algorithm proposed in this paper.

The paper is organized as follows. After introducing our key concepts in Section~\ref{sec:background}, we present in Section~\ref{sec:methods} a Markov Decision Process (MDP) with an extended state space that includes the sequence of goal extracted from the demonstration. This MDP is designed to encourage the agent to achieve the goals only via valid success states. Our approach ensures that the reward received by the agent for reaching a goal is propagated to the valid success states of the previous goals in order to ensure that their value is higher than the value of invalid ones.
To improve the exploration of the state space while training, we transform our MDP into a goal-conditioned MDP (GC-MDP) and adapt the relabelling mechanism of Hindsight Experience Replay (HER) \cite{andrychowicz2017hindsight} to this context. HER relabels the transitions of an episode by replacing the goal initially intended by the agent by the goal it accidentally achieved. Based on this GC-MDP framework, the DCIL-II algorithm is able to efficiently learn a goal-conditioned policy which achieves the full sequence of intermediate goals until task completion.
We then cover related work in Section~\ref{sec:related}. In Section~\ref{sec:experiments}, we evaluate DCIL-II in several challenging robotic simulation scenarios and compare it to state-of-the-art algorithms able to learn complex behaviors by leveraging a single demonstration.

In summary, our main contributions are the following: (1) we present a GCRL framework for learning a control policy to solve complex robotics tasks via sequential goal reaching using a single state-based demonstration; (2) we propose DCIL-II, an instance of this framework which efficiently learns to achieve complex simulated robotics tasks and we show that DCIL-II outperforms state-of-the-art one-demonstration IL methods on these tasks.

\section{Background}
\label{sec:background}

\subsection{(Goal-conditioned) Reinforcement Learning}
\label{sec:GCRL}

To formalize reinforcement learning (RL) problems, one generally calls upon Markov Decision Processes, defined as a tuple $\mathcal{M} = (\mathcal{S}, \mathcal{A}, R, p, \gamma)$. At each step of a time sequence $t=0,1,2,...,T$, the agent selects an action $a_{t}\in \mathcal{A}$ based on its current state $s_{t}\in \mathcal{S}$ using a policy $\pi(a_{t}|s_{t})$. It then moves to a new state $s_{t+1}$ according to an unknown transition probability $p(s_{t+1}|s_{t},a_{t})$ and receives a reward via an unknown reward function $R:\mathcal{S} \times \mathcal{A} \times \mathcal{S} \rightarrow \R$. A discount factor $\gamma$ describes the importance of long-term rewards in this sequence of interactions \cite{sutton1998introduction}. 

The objective in RL is to obtain a policy that maximizes the expected cumulative reward defined as

\begin{equation}
\label{eq:cumulative_reward_RL}
\mathop{\mathbb{E}}_{\pi}[\sum_{t}\gamma^{t}R(s_{t}, a_{t}, s_{t+1})].
\end{equation}

In goal-conditioned reinforcement learning (GCRL) problems \cite{Kaelbling1993learningto, Moore1999multi, schaul2015universal, nasiriany2019planning, chane2021goal}, the MDP framework is extended with a goal space. The reward function is replaced by a goal-conditioned reward function $R:\mathcal{S} \times \mathcal{A} \times \mathcal{S} \times\mathcal{G}\rightarrow \R$ which depends on the considered goal. The goal-conditioned discount function $\gamma:\mathcal{S} \times \mathcal{G} \rightarrow \R$ replaces the discount factor by $0$ when the agent reaches the goal \cite{schaul2015universal}. 

\begin{equation}
\gamma(s,g) = \begin{cases}
0 \text{ if } s=g, \\ 
\gamma \text{ otherwise.}
\end{cases}
\end{equation}

Therefore, all the rewards received after reaching a goal are cancelled. This makes transitions to goal states equivalent to transitions to terminal states, i.e. states terminating the agent trajectory.
The objective in GCRL is to obtain a goal-conditioned policy $\pi(a|s,g)$ which maximizes the expected cumulative reward now defined as  

\begin{equation}
\mathop{\mathbb{E}}_{\pi}[\sum_{t}R(s_{t}, a_{t}, s_{t+1}, g) \prod_{k=0}^{t} \gamma(s_{k}, g)].
\end{equation}

\subsection{Distance-based sparse reward}
\label{sec:distance_based_reward}
In most GCRL problems, the reward function is sparse and the agent only receives a reward for achieving the desired goal. However, in a complex environment with a high-dimensional state space, reaching a goal corresponding to a precise state may be prohibitively difficult, thus it is common to consider goals that represent low-dimensional projections of states \cite{nachum2018data, ecoffet2021first, bagaria2021robustly}. To achieve a goal $g$, the agent must reach any element of the success state set $\mathcal{S}_{g}$, that is any state $s\in \mathcal{S}$ that can be mapped to a goal $g_{s}\in \mathcal{G}$ within a distance less than $\epsilon_{success}$ from $g$, $\epsilon_{success}$ being an environment-dependent hyper-parameter.

In practice, a state is mapped to a goal according to a projection $p_{\mathcal{G}}:\mathcal{S}\rightarrow \mathcal{G}$ associated with the definition of the goal space. In general, the common L2-norm is used to compute the distance between two goals. Besides, the environment-agnostic goal-conditioned reward function is generally defined as

\begin{equation}
    R(s_{t}, a_{t}, s_{t+1},g) = \begin{cases}
1 \text{ if } s_{t+1}\in \mathcal{S}_{g} \\ 
0 \text{ otherwise.}
\end{cases}
\end{equation}

Note that when an agent achieves a goal according to such a distance-based reward, it may have reached a wide variety of success states. This can be problematic if the agent then needs to reach another goal but cannot reach it from certain success states. Thus, the set  of success states $\mathcal{S}_{g}$ for a given goal $g$ can be divided into valid (noted $\mathcal{S}^{valid}_{g}$) or invalid states for reaching the next goal. 

\subsection{Relabelling}

Exploring a high-dimensional state space in a sparse reward context can be challenging for DRL algorithms \cite{houthooft2016vime,bellemare2016unifying,burda2018exploration}. To simplify exploration, GCRL agents often use the Hindsight Experience Replay (HER) relabelling technique \cite{andrychowicz2017hindsight}. During a trajectory, if the agent fails to reach the goal it is conditioned on, HER may relabel some transitions by replacing the initially intended goal with a goal that it happened to achieve. This way, the agent can learn from failed trajectories by receiving rewards for accidentally achieving initially unintended goals. 

\section{Methods}
\label{sec:methods}

In this section we present a general framework where, given a sequence of low-dimensional goals extracted from a single demonstration, an agent must learn to sequentially reach each goal to perform a complex task. We then derive the Divide \& Conquer Imitation Learning II (DCIL-II) algorithm which builds on the framework to obtain such an agent.

\subsection{Problem statement}
\label{sec:problem_formulation}  

DCIL-II is designed to solve a GCRL problem. However, in order to efficiently deal with the sequence of goals extracted from the demonstration, we use a slightly different framework from the one presented in Section~\ref{sec:GCRL}.

\subsubsection{An MDP with implicit sequential goals}
\label{sec:MDPseq}

Given an underlying MDP $\mathcal{M} = (\mathcal{S}, \mathcal{A}, R, p, \gamma)$ with an uninformative reward function $R$ that is non-zero only when the agent reaches a hard-to-reach desired goal, we consider a sequence of low-dimensional goals $\tau_{\mathcal{G}} = \{g_{i}\}_{i\in[0,N_{goals}]}$ leading to the desired goal. 

We extend the state space to include the index of the current goal that the agent must reach. Thus, at step $t$ in a trajectory, an extended state is described by a $(s_{t}, i_{t})$ pair where $i_{t}$ is the index of the current goal. We adapt the reward function, the transition probability distribution and the discount function as follows.  

The index-dependent reward function is defined as

\begin{equation}
\label{eq:reward_seq_MDP}
R_{seq}(\bcancel{s_{t}}, \bcancel{a_{t}}, s_{t+1}, i_{t}) = \begin{cases}
1 \text{ if } s_{t+1} \in \mathcal{S}_{g_{i_{t}}}, \\ 
0 \text{ otherwise.}
\end{cases}
\end{equation}

As shown with the barred $s_{t}$ and $a_{t}$, this function only depends on the next underlying state and the index of the current goal. It rewards the agent when it reaches the success states associated with its current goal.

The agent must reach the successive goals sequentially. There might be different approaches to deal with this sequentiality constraint, but in the implementation presented here, the index is automatically incremented after a successful transition:

\begin{equation}
\label{eq:index_seq_MDP}
i_{t+1} = \begin{cases}
i_{t} + 1 \text{ if }  s_{t+1} \in \mathcal{S}_{g_{i_{t}}}, \\ 
i_{t} \text{ otherwise.}
\end{cases}
\end{equation}

These automatic switches of indices are included in the joint transition probability which is defined as

\begin{equation}
\begin{split}
p_{seq}(s_{t+1}, i_{t+1}|s_{t}, a_{t}, i_{t}) \\
= p(i_{t+1}|s_{t+1}, &\bcancel{s_{t}}, \bcancel{a_{t}}, i_{t})p(s_{t+1}|s_{t}, a_{t}, \bcancel{i_{t}}) \\
= p(i_{t+1}|s_{t+1}, &i_{t})p(s_{t+1}|s_{t}, a_{t}),
\end{split}
\end{equation}

with $p(s_{t+1} | s_{t}, a_{t})$, the index-independent state transition probability of the underlying MDP and $p(i_{t+1} | s_{t+1}, i_{t})$ the index transition probability. This latter corresponds to a deterministic distribution as the index at step $t+1$ is a constant random variable when conditioned on the next state and the current index. It can be defined as follows: 

\begin{equation}
\label{eq:joint_dist_transition}
\begin{split}
p(i_{t}+1|s_{t+1}, i_{t}) = R_{seq}(s_{t+1},i_{t}),\\
p(i_{t}|s_{t+1}, i_{t}) = 1-R_{seq}(s_{t+1},i_{t}).\\
\end{split}
\end{equation}

When the agent reaches the final goal of the sequence, it enters a terminal state. 
So we define the discount function as

\begin{equation}
\label{eq:gamma_seq}
\gamma_{seq}(s) = \begin{cases}
0 \text{ if } s \in \mathcal{S}_{g_{N_{goal}}}, \\ 
\gamma \text{ otherwise.}
\end{cases}
\end{equation}

\begin{figure}[t!]
     \centering
     \includegraphics[width = \hsize]{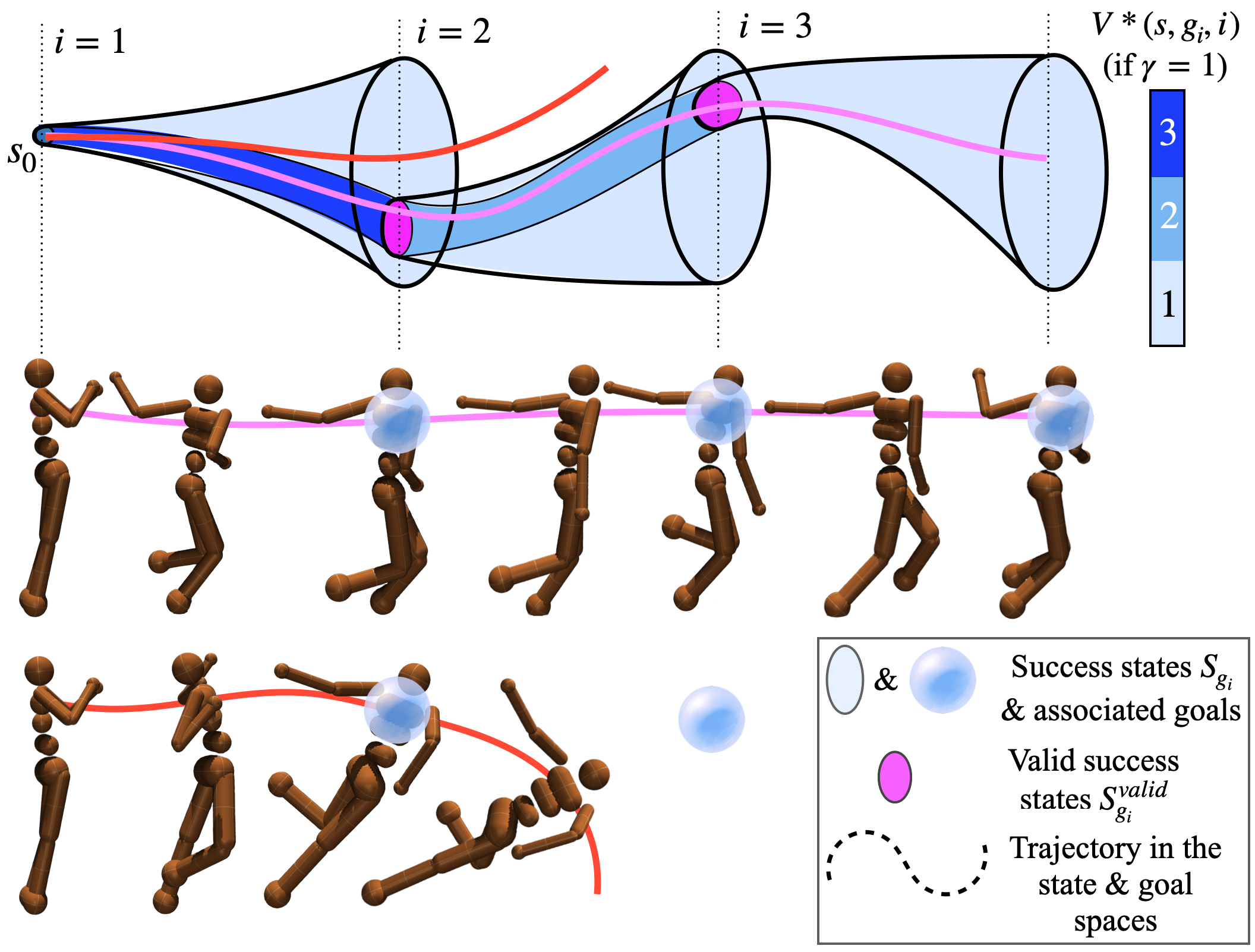}
    \caption{To successively reach each goal, the agent must transit between successive sets of valid success states. The non-terminal success states propagate the reward from each index $i$ to the previous ones. \textbf{Top:} value propagation in a simplified 2D representation of the state space when $\gamma=1$. In the shown environment, goals correspond to successive Cartesian positions of the torso of a humanoid. \textbf{Middle:} trajectory where the agent successfully transitions from one valid success state set to another. \textbf{Bottom:} trajectory where the agent jumps to achieve the first goal but does it by reaching a invalid success state which prevents it from reaching the next goal.}
    \label{fig:valid_invalid_success_states} 
\end{figure}

This function makes the success states associated with the intermediate goals in $\tau_{\mathcal{G}}$ non-terminal except for the final goal. 

As discussed in Section~\ref{sec:distance_based_reward}, reaching a goal corresponding to a low dimensional projection of a state does not fully condition the state of the agent. In particular, if some of the success states are invalid, the agent must avoid them. 

The definition of the discount function in \eqref{eq:gamma_seq} implicitly encourages the agent to reach each goal by only reaching valid success states. Indeed, non-terminal success states associated with intermediate goals in $\tau_{\mathcal{G}}$ are used to propagate the value of the next goal to the previous one via the value approximation $\Tilde{V}$ of successful transitions. Using \eqref{eq:reward_seq_MDP} and \eqref{eq:gamma_seq} we get

\begin{equation}
\begin{split}
\label{eq:reward_propagation}
\Tilde{V}(s_{t}, i_{t}) &= R_{seq}(s_{t}, a_{t}, s_{t+1},i_{t}) + \gamma_{seq}(s_{t+1}) V^{\pi}(s_{t+1}, i_{t}+1)\\
&= 1 + \gamma V^{\pi}(s_{t+1}, i_{t}+1).
\end{split}
\end{equation}

After enough training, the value $V^{\pi}(., i_{t}+1)$ of valid success states associated with the next index should be larger than the value of invalid ones as the agent was only rewarded for trajectories starting from the former and not from the latter. 

Therefore, both valid and invalid success states receive the sparse reward for reaching the current goal, but valid ones benefit from the propagation of a larger value from the next state. This difference is illustrated in \figurename~\ref{fig:valid_invalid_success_states}.

Using the extended state space, adapted reward function, transition probability and discount function defined above, we end up with an alternative MDP $\mathcal{M}_{seq} = (\mathcal{S}_{seq}, \mathcal{A}, R_{seq}, p_{seq}, \gamma_{seq})$. In this MDP, we condition the policy implicitly on the goals of the sequence by using their indices in the extended state space. 

The objective in $\mathcal{M}_{seq}$ is to obtain a policy that maximizes the expected cumulative rewards:

\begin{equation}
\label{eq:obj_MDP_seq}
\mathop{\mathbb{E}}_{\pi}[\sum_{t}R_{seq}(s_{t+1}, i_{t})\prod_{k=0}^{t}\gamma_{seq}(s_{k})].
\end{equation}

Maximizing this equation boils down to reaching each goal in $\tau_{\mathcal{G}}$ by passing through each set of valid success states. 

Nevertheless, learning how to reach each goal using a sparse reward like $R_{seq}$ can be challenging. Moving to GCRL, we can address this issue by leveraging the relabeling mechanism of HER.

\subsubsection{Moving to GCRL to improve exploration}
\label{sec:GCMDP}

In Section~\ref{sec:MDPseq}, we defined a traditional MDP with an extended state. We now call upon the GCRL framework. In addition to implicit goal-conditioning via the goal indices, we now also explicitly condition the policy $\pi(a|s, i, g_{i})$ on the current goal $g_{i}$. As a result, we can use a relabelling mechanism like HER to reward transitions towards unintended goals and improve the exploration of the agent while it is training. 

\begin{figure}[b!]
     \centering
     \includegraphics[width = \hsize]{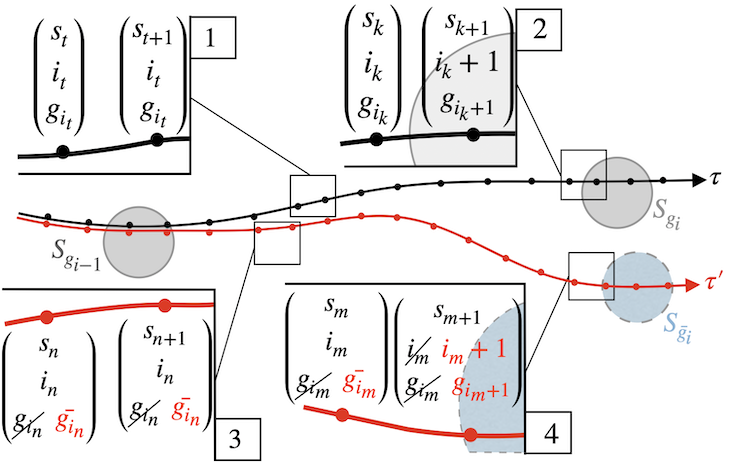}
    \caption{Automatic switches of goals and indices. When the agent reaches its current goal $g_{i}$ as in the black trajectory, the current index $i$ is incremented and the goal is switched to the next one in the sequence $\boxed{2}$. When the agent fails to reach the current goal or relabeled goal $\bar{g_{i}}$, the goal and the index keep the same $\boxed{1}$ \& $\boxed{3}$. For transitions relabeled as successful as in the red trajectory with a relabeled goal $\bar{g_{i}}$, the current and next indices and goals are relabeled to include these automatic switches $\boxed{4}$.}
    \label{fig:transitions_DCILII} 
\end{figure}

As with HER, during policy updates, a failed transition can be artificially relabeled as successful and the missed goal can be replaced by an unintended goal achieved later in the trajectory. 
However, unlike traditional GCRL, the goal changes along the trajectory as the agent must reach each goal sequentially. To include the automatic goal switches defined in \eqref{eq:index_seq_MDP} and \eqref{eq:joint_dist_transition} and let the value  propagate between successive goals as in \eqref{eq:reward_propagation}, we must relabel the next goal and next index too. 
Thus, in a failed transition relabeled as successful, the original next goal is replaced by the next goal in the sequence according to the current index. Besides, the next index corresponds to the incremented current index (see \figurename~\ref{fig:transitions_DCILII}). 

To deal appropriately with relabelling in this context, we distinguish four different types of transitions $(s_{t}, i_{t}, g_{t})\rightarrow_{a_{t}} (s_{t+1}, i_{t+1}, g_{t+1})$: original failed transitions $\boxed{1}$, original successful transitions $\boxed{2}$, transitions relabeled as failed  for artificial goal $\bar{g_{t}}$ $\boxed{3}$ or transitions relabeled as successful for artificial goal $\bar{g_{t}}$ $\boxed{4}$. For each type of transition, the current and next goals and indices vary according to 

\begin{equation}
\label{eq:next_goal_next_index}
(g_{t+1}, i_{t+1}) = f_{\text{next goal \& index}}(s_{t+1}, g_{t}, i_{t}) 
\end{equation}

with the $f_{\text{next goal \& index}}:S\times G \times \mathbb{N} \rightarrow G \times \mathbb{N}$ function returning the next goal and index given the current ones and the next state. This function switches to the next goal in the sequence according to the current index when the current goal is reached: 

\begin{equation}
f_{\text{next goal \& index}}(s_{t+1}, g_{t}, i_{t}) := \begin{cases}
(g_{i_{t}+1}, i_{t}+1) \text{ if } s_{t+1} \in \mathcal{S}_{g_{i_{t}}} \\
(g_{i_{t}}, i_{t}) \text{ if } s_{t+1} \cancel{\in} \mathcal{S}_{g_{i_{t}}}.
\end{cases}
\end{equation}

Using $f_{\text{next goal \& index}}$ to replace the next goal and index in both original and relabeled transitions, one can leverage HER in sparse reward contexts.

In addition, we also get value approximations similar to \eqref{eq:reward_propagation} for original successful transitions $\boxed{2}$:

\begin{equation}
\label{eq:reward_propagation_GCRL_2}
\Tilde{V}(s_{t}, i_{t}, g_{i_{t}}) = 1 + \gamma V^{\pi}(s_{t+1}, i_{t}+1, g_{i_{t}+1})
\end{equation}

and transitions relabeled as successful for artificial goal $\bar{g}$ $\boxed{4}$:

\begin{equation}
\label{eq:reward_propagation_GCRL_4}
\Tilde{V}(s_{t}, i_{t}, \bar{g}) = 1 + \gamma V^{\pi}(s_{t+1}, i_{t}+1, g_{i_{t}+1})
\end{equation}

Thus, both types of transitions propagate the value associated with the next goal $V(., i_{t}+1,.)$ in $\tau_{\mathcal{G}}$ into the value associated with the current one $V(., i_{t}, .)$.
Therefore, whatever the goal $g$ conditioning the policy $\pi(a|s, i, g)$, the agent is always encouraged to prepare for the next goals in $\tau_{\mathcal{G}}$ while reaching the current one.

Note that if we decided to remove the index in the state to recover the usual GCRL state space (see Section~\ref{sec:GCRL}) and to condition the policy $\pi(a|s, g)$ on the explicit goal only, we could not use a similar reward propagation mechanism. 

Indeed, let's consider two failed training trajectories where the agent aimed for two different goals $(g_{i},g_{j})$ in $\tau_{\mathcal{G}}$ and transited by the same states $(s_{t},s_{t+1})$. In a transition relabeled as successful for artificial goal $\bar{g}=p_{\mathcal{G}}(s_{t+1})$, the relabeled next goals $(g_{i+1},g_{j+1})$, defined according to the position of original goals in $\tau_{\mathcal{G}}$, are different. However, without any index, the states and the relabeled current goals $(\bar{g}_{i},\bar{g}_{j})=(\bar{g},\bar{g})$ would be the same.

In this context, the value approximation $\Tilde{V}$ of the same state-goal pair $(s_{t}, \bar{g})$ would be approximated differently depending on the chosen transition as the value of the next state would be conditioned differently.

For the transition originally aiming for $g_{i}$, we would have
\begin{equation}
\begin{split}
\Tilde{V}(s_{t}, \bar{g}) &= R_{seq}(s_{t}, a_{t}, s_{t+1},\bar{g}) + \gamma_{seq}(s_{t+1}) V^{\pi}(s_{t+1}, g_{i+1}),\\
&= 1 + \gamma V^{\pi}(s_{t+1}, \underline{g_{i+1}}),
\end{split}
\end{equation}
\noindent
whereas for the transition originally aiming for $g_{j}$, we would have
\begin{equation}
\begin{split}
\Tilde{V}(s_{t}, \bar{g}) &= R_{seq}(s_{t}, a_{t}, s_{t+1},\bar{g}) + \gamma_{seq}(s_{t+1}) V^{\pi}(s_{t+1}, g_{j+1}),\\
&= 1 + \gamma V^{\pi}(s_{t+1}, \underline{g_{j+1}}).\\
\end{split}
\end{equation}

Therefore, using our relabelling mechanism for reward propagation between goals in a GC-MDP without index would lead to a partially observable MDP as the agent would not be aware of which goal in $\tau_{\mathcal{G}}$ it is targeting, which would result in unstable value learning \cite{sutton1998introduction}.

These properties are demonstrated experimentally in \figurename~\ref{fig:comp_value_propagation} with ablations described in Section~\ref{sec:ablation_study}.

\begin{algorithm}[t!]
\caption{DCIL-II}
\label{algo:GCP_training}
\begin{algorithmic}[1]
    \State \textbf{Input:} $\pi, Q, \bar{Q}, \tau_{demo}$ 
    \Comment{actor/critic/target critic networks \& demonstration}
    \State $\{g_{i}\}_{i\in[1,N_{goals}]} \leftarrow$ extract\_goals$(\tau_{demo})$
    \State $B\leftarrow [\ ]$ 
    \Comment{replay-buffer}
    \State $R \leftarrow \{i:[\ ]\}_{i\in[1,N_{goals}]}$
    \Comment{successes/failures memory}
    \For{$n=1:N_{episode}$}
        \State $ i \leftarrow select\_index(R)$
        \Comment{\textbf{Trajectory initialization}} \label{lst:line:begin_traj_init}
        \State $s \leftarrow \text{env.reset}(i)$
        \State $success, done, last\_index \leftarrow False, False, False$ \label{lst:line:end_traj_init}
        \While{not done} 
        \Comment{\textbf{Trajectory rollout}} \label{lst:line:begin_traj_rollout}
            \State $a \sim \pi(a|s,i,g_{i})$
            \State $s', env\_done \leftarrow env.step(a)$ 
            \State $r \leftarrow 0$
            \If{$s' \in \mathcal{S}_{g_{i}}$}
            \Comment{success}
                \State $r, success \leftarrow 1, True$
                \State $last\_index \leftarrow (i \geq nb\_skills)$
                \State $i' \leftarrow i+1$
                \Comment{index shift}
                \State $R[i] \leftarrow R[i] + [1]$                \label{lst:line:tack_success}
            \ElsIf{$t \geq  T_{max}\ \text{or}\ env\_done$}
                \Comment{failure}
                \State $success, timeout \leftarrow False, True$
                \State $i' \leftarrow i$
                \State $R[i] \leftarrow R[i] + [0]$                \label{lst:line:tack_failure_1}
            \Else
                \State $success, timeout \leftarrow False, False$
                \State $i' \leftarrow i$
            \EndIf
            \If{$env\_done\ \text{or}\ last\_index$}
                \State $done \leftarrow True$
            \label{lst:line:tack_failure_2}
            \EndIf
            \State $B \leftarrow B + (s, g_{i}, i, a, s', g_{i'}, i', r, done, success)$
            \State $s,i \leftarrow s',i'$
            \State $done \leftarrow done \lor timeout$ \label{lst:line:end_traj_rollout}
            \State $\text{SAC\_update}(\pi, Q, \bar{Q},B)$ \label{lst:line:sac_update}
        \EndWhile
    \EndFor
\end{algorithmic}
\end{algorithm}

\subsection{The DCIL-II algorithm}
\label{sec:dcil2_algo}

The DCIL-II algorithm first extracts the sequence of goals from a single demonstration and derives the alternative GC-MDP introduced in Section~\ref{sec:GCMDP}. By running a 2-step loop, it then learns a policy that can be used later to sequentially reach each goal and complete the complex demonstrated behavior. Algorithm \ref{algo:GCP_training} summarizes these different steps.

\subsubsection{From a single demonstration to a sequence of goals}

To extract the sequence of goals $\tau_{\mathcal{G}}$ from the demonstration, we project the demonstrated states in the goal space and split the demonstration into $N_{goal}$ sub-trajectories of equal arc lengths $\epsilon_{\text{dist}}$ as in \cite{Chenu2022divide}. For each sub-trajectory in the goal space, we extract its final elements and concatenate them to construct $\tau_{\mathcal{G}}$. We also extract the states associated with the initial elements of each sub-trajectories. Those states form a set of demonstrated states used by DCIL-II to reset the agent in a valid success state and to train it at reaching the following goal (Hypothesis~\ref{hyp:reset}).

\subsubsection{Main Loop}

DCIL-II runs a 2-step loop to train an agent and combines an off-policy actor-critic algorithm (e.g. the Soft Actor-Critic (SAC) algorithm \cite{haarnoja2018soft}) with the HER-like relabelling mechanism described in Section \ref{sec:GCMDP} to learn the goal-conditioned policy.  

\paragraph{Trajectory initialization (lines~\ref{lst:line:begin_traj_init}-\ref{lst:line:end_traj_init})} DCIL-II selects an index $i$ in the sequence $[1,N_{goal}]$ and the agent is reset in the associated demonstrated state. 

\paragraph{Trajectory rollout (lines~\ref{lst:line:begin_traj_rollout}-\ref{lst:line:end_traj_rollout})} This reset state is extended with the selected index. The policy is conditioned on the state, the selected index $i$ and the associated goal $g_{i}$. The agent then starts a trajectory that is terminated either if the agent reaches each goal successively up to the final one, if a time limit is reached or if the agent reaches a terminal state. 

This loop is repeated after the termination of each trajectory. DCIL-II keeps track of the successes and failures for each indexed trajectory (lines \ref{lst:line:tack_success}, \ref{lst:line:tack_failure_1} and \ref{lst:line:tack_failure_2}). Thus, the index selection (line~\ref{lst:line:begin_traj_init}) can be biased towards indices corresponding to goals with a low ratio of success using a discrete distribution over indices that is inverse proportional to the success ratio. Such distribution is obtained using the roulette wheel selection operator commonly used in Evolutionary Algorithms \cite{Blickle1996selection}.

The saved transitions are used to perform an actor-critic update after each step in the environment (line~\ref{lst:line:sac_update}). In a sampled batch of transitions, half of the transitions are relabeled using the relabelling mechanism presented in Section \ref{sec:GCMDP}.

\section{Related Work}
\label{sec:related}
Our method performs imitation learning by leveraging a single demonstration, it leverages a sequential goal-reaching bias, and it focuses on reaching valid success states only using a value propagation mechanism. We cover the literature in these three topics taken separately, then we pay more attention to three works which include mechanisms close to those of DCIL-II. These three works will inspire the three ablations considered in Section~\ref{sec:DCIL_i_DCIL_ii}. Table~\ref{tab:hypothesis} summarizes the similarities and differences between our method and the main ones mentioned in this Section.


\subsection{Learning from a single demonstration}
Only a few works have focused on leveraging a single state-based demonstration to learn a policy to solve complex tasks. IL methods based on Behavioral Cloning (BC) \cite{pomerleau1991efficient} or Inverse Reinforcement Learning (IRL) \cite{russell1998learning} fail to imitate complex behaviors when a single demonstration is available \citep{pmlr-v9-ross10a, resnick2018backplay, behbahani2019learning}.
Backplay \cite{resnick2018backplay, salimans2018backplay} and Primal Wasserstein Imitation Learning (PWIL) \cite{dadashi2020primal} are two DRL-based methods designed to imitate complex behaviors from a single demonstration. 
Backplay has been successfully applied to play Atari games and to learn a control policy for a robotic grasping task using a single demonstration \cite{ecoffet2019go, ecoffet2021first}. However, the method often suffers from catastrophic forgetting \cite{matheron2020pbcs} which reduces its applicability to complex imitation tasks. 
PWIL is a recent IL algorithm which minimizes a greedy version of the Wasserstein distance between state-action distributions of the expert and the agent. PWIL was the first IL method to successfully learn a control policy for a locomotion task with a simulated humanoid using a single demonstration. However, it does not leverage sequential goal reaching in contrast with DCIL-II.

\subsection{Sequential goal reaching}
To imitate from a single demonstration, DCIL-II relies on sequential goal achievement, which is the central tool of several other RL methods. 

In Feudal Hierarchical Reinforcement Learning (HRL) methods \cite{dayan1992feudal,nachum2018data,Levy2019LearningMH}, a high-level policy constructs a sequence of goals on-the-fly by periodically assigning goals to a low-level goal-conditioned policy over the entire episode. However, jointly training a multi-stage policy induces training instabilities caused by non-stationary transitions function \cite{nachum2018data, hutsebaut2022hierarchical}.

To achieve a difficult to reach goal, several approaches use the goal-conditioned value function of the agent to obtain a sequence of simpler goals \cite{nair2018visual, chane2021goal, nasiriany2019planning, eysenbach2019search}. However, these methods rely on an accurate value function which is rarely available in hard-exploration problems. Their applicability either assumes that the agent can easily explore the entire state space or that the agent can be reset in uniformly sampled states. 
By contrast, we assume hard exploration environments with a sparse reward function and do not leverage a reset-anywhere property, as DCIL-II only resets along the demonstrated trajectory.

The policy-based version of the Go-Explore algorithm \cite{ecoffet2021first} also guides the agent with a sequence of goals
extracted from training trajectories.
These goals are used to guide the agent back to previously visited states in order to continue the exploration process directly from advanced states. However, the authors did not dwell on the problem of invalid success states when reaching successive goals. In our method, we tackle this problem using value propagation, a mechanism also used in Option-based HRL, as described below.

\subsection{Value propagation mechanisms}

In Option-based HRL \cite{sutton1999between, precup2000temporal}, an option contains: 1) a set of initial states where the option can be activated, 2) a termination function  deciding if the option should be terminated given the current state, 3) the intra-option policy which controls the agent at a single time step scale and, 4) the policy over options which decides which option to trigger when the agent completed the current option. 

When intra-option policies are conditioned on goals, HRL relies on sequential goal-reaching. Indeed, similarly to Feudal HRL, a sequence of goals is built on-the-fly by the policy over options. This view was adopted in Robust Deep Skill Chaining (R-DSC) \cite{bagaria2021robustly} which trains goal-conditioned intra-option policies using HER.
In R-DSC, the value of an option is propagated to the value of the previous option in order to select the optimal goal for chaining these two successive options.

Similar to our framework, other option-based HRL methods \cite{levy2011unified, bacon2017option} designed an MDP with an extended state space to facilitate switches between options by propagating the value from one to another. However, this framework has not yet been adapted to goal-conditioned policies. Moreover, these option-based HRL methods as well as R-DSC do not learn from demonstrations.

Closest to our work, the first version of the DCIL algorithm \cite{Chenu2022divide} (now referred to as DCIL-I) uses a reward bonus and an overshoot mechanism to propagate the value of a goal to the value of the previous one in a sequence of goals similar to $\tau_{\mathcal{G}}$. DCIL-I is also designed to imitate complex behaviors using a single demonstration. Moreover, it also tackles the problem of reaching valid success states only while leveraging a sequential goal-reaching bias.

In the following paragraphs, we discuss the relations and differences between the value propagation mechanisms used in DCIL-I, option-based HRL, R-DSC and ours.

\subsection{Bridging the gap between DCIL-I \& DCIL-II}
\label{sec:DCIL_i_DCIL_ii}

\begin{figure*}[thbp]
     \centering
     \includegraphics[width=\hsize]{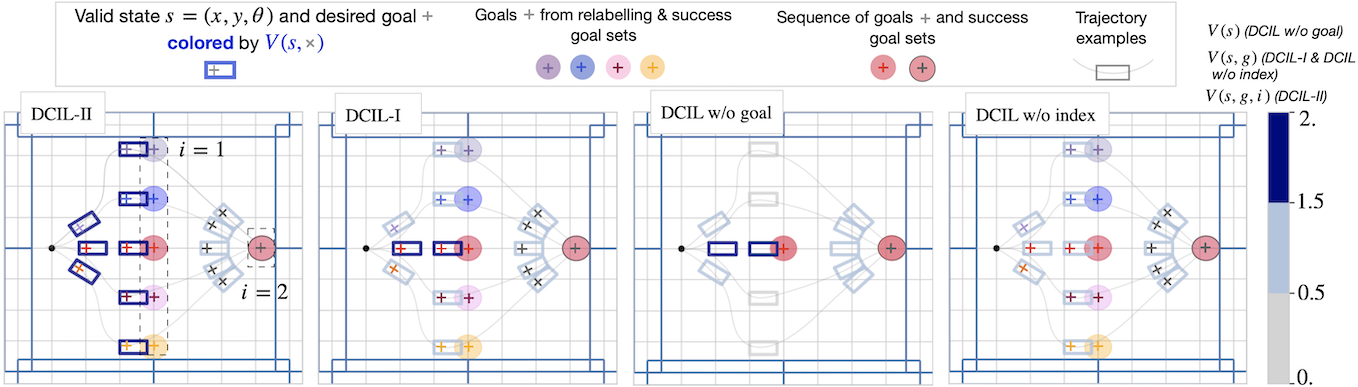}
    \caption{Comparison of value propagation in DCIL-II, DCIL-I, DCIL w/o goal and DCIL w/o index, using a Dubins car environment after 15.000 training steps. Unlike when success states are terminal as in DCIL w/o index, the value propagation mechanisms in DCIL-II, DCIL-I and DCIL w/o goal help propagate the value of the second goal into the value of the first ($(s,g_{i})$ pairs with a value $\sim 2$, in dark blue).  Moreover, the relabelling mechanism of DCIL-II helps propagate the value associated with the second goal to the value associated with any relabeled goal (relabeled $(s,\bar{g}_{i})$ pairs with a value $\sim 2$). In DCIL-I, the same value is propagated to the first goal in $\tau_{\mathcal{G}}$ only (relabeled $(s,\bar{g}_{i})$ pairs with a value $\sim 1$, in light blue). Those differences between value propagation mechanisms result in the larger value of valid states leading to alternative intermediate goals in DCIL-II compared to any other methods. In other words, DCIL-II helps the agent prepare for the next goal while reaching any intermediate goal.}
    \label{fig:comp_value_propagation} 
\end{figure*}

In DCIL-I, instead of considering a single MDP with an extended state space, the authors rather considered a sequence of MDPs. In this sequence of MDPs $\mathcal{M}_{i} = (\mathcal{S}, \mathcal{A}, R_{i}, p, \gamma_{i})$, only the reward function and the discount function differed from one MDP to another. First, the discount function made the success states associated to any goal terminal:

\begin{equation}
\gamma_{i}(s) = \begin{cases}
0 \text{ if } s \in \mathcal{S}_{g_{i}}, \\ 
\gamma \text{ otherwise.}
\end{cases}
\end{equation}

In addition, the sparse reward received by the agent when it reached the current goal was augmented by a \textit{reward bonus} based on the value with respect to the next MDP $V_{i+1}(s)$:

\begin{equation}
\label{eq:reward_MDP_i}
R_{i}(s_{t+1}) = \begin{cases}
1 + V_{i+1}(s_{t+1}) \text{ if } s_{t+1} \in \mathcal{S}_{g_{i}}, \\ 
0 \text{ otherwise.}
\end{cases}
\end{equation}

This reward bonus was used to propagate the value from the next MDPs to the previous ones and was analogous to the non-terminal success states in DCIL-II (see Section~\ref{sec:problem_formulation}). Indeed, it encouraged the agent to achieve the goal of the current MDP by reaching valid success states that were compatible with the resolution of the next MDPs. Thus, the optimal policy for one MDP in the sequence depended on the next MDP. 

Using this sequence of MDPs instead of a single MDP, the authors avoided extending the state space to include the index of the current goal. However, as illustrated in \figurename~\ref{fig:comp_value_propagation}, when this framework was combined with HER, the agent only received the reward bonus when it reached the goals of the sequence but not when it reached the artificial goals created by the relabelling process. Thus, the agent was encouraged to prepare for the sequence of goals only when it was conditioned on the goals of the sequence. On the contrary, in the single MDP framework used by DCIL-II, the extended state space combined with the modified relabelling mechanism forces the agent to prepare for the next goals in the sequence while reaching both types of goals (goals of the sequence or relabeled goals). 

We further underline the difference between the value propagation mechanisms of DCIL-I and DCIL-II in the ablation study conducted in Section~\ref{sec:ablation_study}.

\subsection{Bridging the gap between DCIL-II \& option-based HRL}
\label{sec:DCIL_ii_HRL}

We can look at the MDP $\mathcal{M}_{seq}$ (Section~\ref{sec:problem_formulation}) from the perspective of the Option-based HRL framework \cite{precup2000temporal} and see it as a semi-MDP with a hand-designed chain of options.


Indeed, we can associate an option to each goal in $\tau_{\mathcal{G}}$ by considering the set of success states of these goals as termination sets and the policy conditioned on the indices of these goals as the intra-option policies. From this perspective, we get a chain of options by considering that the termination set associated with a goal is the initiation set of the option associated with the next goal, as in \cite{bagaria2019option}. Thus, we also have a fixed and deterministic policy over options switching to the option associated with the next goal in $\tau_{\mathcal{G}}$ as soon as the current option is completed.

We thus see that $\mathcal{M}_{seq}$ defines a restricted semi-MDP where the structure of options is a simple chain and an option can only trigger the next one. 

Nevertheless, in this semi-MDP, values are propagated from one option in the chain to the previous ones via non-terminal success states. A similar semi-MDP which also propagates value between options is defined in the Option-Critic framework \cite{bacon2017option, klissarov2017learnings, harb2018waiting}. However, the two-stage learning of both the policy over options and the intra-option policies makes the end-to-end training of those policies unstable \cite{hutsebaut2022hierarchical}.
Moreover, in the absence of explicit goal-conditioning, no relabelling mechanism may be applied in $\mathcal{M}_{seq}$ or in any method based on the Option-Critic framework. This makes exploration difficult when learning how to reach sparsely rewarded goals.

By extending $\mathcal{M}_{seq}$ to GCRL (Section~\ref{sec:GCMDP}), DCIL-II can be seen as a modified Option-based method which benefits from both reward propagation and exploration mechanisms at the cost of hand-designing options. 

Again, we underline the importance of combining the value propagation mechanism with an exploration mechanism by considering an ablation of DCIL-II without HER called {\em DCIL w/o goal} in Section~\ref{sec:ablation_study}.

\subsection{Bridging the gap between DCIL-II \& goal-conditionned skill-chaining}
\label{sec:DCIL_i_RDSC}

In DSC \cite{bagaria2019option}, the authors construct a chain of options backward from the goal. In R-DSC \cite{bagaria2021robustly} which is a refined version of DSC, goal-conditioned policies are used as intra-option policies and HER is used to train them. This capability to benefit from HER makes R-DSC very close to DCIL-II. 
However, in R-DSC, options are not hand-designed using a sequence of goals. Instead, a learned classifier delimits the termination set of each option and a goal is sampled within this non-stationary set of states. 

For each option, a set of states from which the agent has been able to achieve the termination set of the next option is constructed and updated after each episode. Within the set of goals obtained by projecting these states in the goal space, the one with the largest value is selected to condition the intra-option policy. 
The values of those goals are updated after each episode using the following relation:

\begin{equation}
\begin{split}
V_{g}(p_{\mathcal{G}}(s),&p_{\mathcal{G}}(s')) =\\
&V_{o}(s, p_{\mathcal{G}}(s')) +  \gamma \max_{s''\in \epsilon_{p}}V_{g}(p_{\mathcal{G}}(s'), p_{\mathcal{G}}(s''))
\end{split}
\end{equation}
\noindent
where $V_{g}$ returns the value of goals at the level of options and $V_{o}$ returns the value of states at the level of control steps.

Thus, the value is only propagated from one option to the others for the selection of goals and does not impact intra-option policies. Therefore, it is implicitly assumed that the trained intra-option aiming for the selected goal will only reach valid success states. 
However, if the intra-option policy changes and eventually leads the agent to invalid success states, the agent is not able to complete each option successively. Therefore, the option chain is broken until selecting another goal reached via valid success states only.


By contrast, the DCIL-II mechanism for value propagation avoids these breaks by increasing the value of valid success states and, therefore, encouraging the agent to reach each goal only via these states.

As before, we underline the importance of this value propagation mechanism by comparing DCIL-II to an ablated version which uses the same vanilla HER as R-DSC called {\em DCIL w/o index} (see Section~\ref{sec:ablation_study}).

\subsection{Comparison of hypotheses}
\label{sec:hypothesis}

The DCIL-II algorithm relies on three main hypotheses. We assume that the agent can be reset in some states of the demonstration, that expert actions are not provided and that a definition of the goal space is given to perform GCRL. These hypotheses are summarized in Table~\ref{tab:hypothesis}.

\subsubsection{Reset}
\label{hyp:reset} The training procedure in DCIL-II assumes that the agent can be reset in some demonstrated states. A similar form of reset is assumed in DCIL-I. Stronger reset assumptions like \textit{reset-anywhere} where the agent can be reset in uniformly sampled states have also been considered in the GCRL literature \cite{nasiriany2019planning}. 
PWIL is based on the more classical assumption of a unique reset, and BC does not require any reset at all.

\subsubsection{State-only demonstration}
\label{hyp:state_only}
Similarly to DCIL-I, DCIL-II learns the desired complex behahior by leveraging a single state-based demonstration, without considering the actions. Learning from states only is crucial when the actions used in the demonstration are difficult to collect (e.g. motion capture, human guidance). On the contrary, both PWIL and BC require state-action demonstrations. 

\subsubsection{Goal-space definition}
\label{hyp:goal_space}
As GCRL-based methods, DCIL-II and DCIL-I require a definition of the goal space and the corresponding mapping from the state space to the goal space. No such assumption is necessary in PWIL or BC as none of them are based on GCRL. 

A summary of these hypotheses has been included in Table~\ref{tab:hypothesis}.

\begin{table*}[t]
\centering
\begin{tabular}{ |M{3cm}|M{1cm}M{1cm}M{1cm}M{1cm}M{1cm}M{1cm}|}
 \hline
  & \textit{DCIL-II} & \textit{DCIL-I} & \textit{PWIL} & \textit{BC} & \textit{R-DSC} & \textit{Option-Critic} 
  \\
\hline
\rowcolor{MediumGray}
 Learning from demonstration  & $\bullet$ & $\bullet$ & $\bullet$ & $\bullet$ &  &  
 \\
 Sequential goal-reaching bias & $\bullet$ & $\bullet$ & & & $\bullet$ &   
 \\
 Relabelling mechanism & $\bullet$ & $\bullet$ & & & $\bullet$ & 
 \\
 Value propagation mechanism & $\bullet$ & $\bullet$ & & & $\bullet$ & $\bullet$ 
 \\
 Extended state space & $\bullet$ & & & & & $\bullet$ 
 \\
 \rowcolor{DarkGray}
 Reset hypothesis: & & & & & & 
 \\ 
 \rowcolor{DarkGray}
 $s_{0}$ - single state  & $\{s_{i}\}$ & $\{s_{i}\}$ & $s_{0}$ &  & $s_{0}$ & $s_{0}$ 
 \\ 
 \rowcolor{DarkGray}
 $\{s_{i}\}$ - states from demo & & & & & & 
 \\
 \rowcolor{DarkGray}
  Demonstration type: & & & & & & 
 \\
 \rowcolor{DarkGray}
 $S\times A$ - states and action & $S$ & $S$ & $S\times A$ & $S\times A$ &  & 
 \\
 \rowcolor{DarkGray}
 $S$ - states only & & & & & & 
 \\
 \rowcolor{DarkGray}
 Goal space definition & $\bullet$ & $\bullet$ & & & $\bullet$ & $\bullet$ 
 \\
 \hline
\end{tabular}
\caption{Comparison of the problems tackled (dark grey), the mechanisms used (white) and the hypothesis made (light grey) by DCIL-II, three methods tackling the same problem of learning from a single demonstration (DCIL-I \cite{Chenu2022divide}, PWIL \cite{dadashi2020primal} and BC \cite{pomerleau1991efficient}) and two methods relying  on a value propagation mechanism (R-DSC \cite{bagaria2021robustly} and Option-Critic \cite{bacon2017option}).}
\label{tab:hypothesis}
\end{table*}


\section{Experiments}
\label{sec:experiments}
In this section, we first introduce the experimental setup consisting of five environments of varying complexity. We then present the implementation details of DCIL-II and conduct an ablation study to empirically validate our design choices. Finally, we compare our method to different baselines and apply it to a more realistic simulated robotic environment.

\subsection{Environments}

We evaluate DCIL-II in five environments: the \textit{Dubins Maze} environment \cite{Chenu2022divide}, the \textit{Fetch} environment \cite{ecoffet2021first}, two variants of the \textit{Humanoid} environment from the OpenAI Gym Mujoco suite \cite{Todorov2012mujoco} \cite{brockman2016openai} and the \textit{Cassie Run} environment based on \cite{menagerie2022github}. 

\subsubsection{Dubins Maze}

The \textit{Dubins Maze} is a navigation task where the agent controls a Dubins car \cite{dubins1957curves} in a 2D maze. The state $s=(x,y,\theta) \in X\times Y\times \Theta$ includes the 2D position of the car in the maze and its orientation. The forward veolcity being constant, the agent only controls the variation of orientation $\dot{\theta}\in \mathbb{R}$ of the car. The goal space is defined as $X\times Y$, thus goals correspond to 2D positions. Such goal space design does not condition the orientation of the car when the agent reaches a goal. Demonstrations are obtained using the Rapidly-Exploring Random Trees (RRT) algorithm \cite{lavalle1998rapidly}.    

\subsubsection{Fetch}

The \textit{Fetch} environment is a simulated grasping task for a 8 degrees-of-freedom robot manipulator. A sparse reward is obtained only when the agent grasped an object and put it on a shelf. 
The state $s\in \mathbb{R}^{604}$ contains the Cartesian and angular positions and the velocity of each
element in the environment (robot, object, shelf, doors...) as well as the contact Boolean evaluated for each pair of elements. 
In this environment, a goal $g\in \mathbb{R}^{6}$ corresponds to the concatenation of the Cartesian position of the end-effector of the robot and the object. Therefore, the agent may reach a goal with an invalid orientation or velocity that may prevent grasping. 
Demonstrations are obtained using the exploration phase of the Go-Explore algorithm \cite{ecoffet2021first}.

\subsubsection{Humanoid locomotion}
\label{sec:humanoid_locomotion}

The \textit{Humanoid locomotion} environment is a variant of the Humanoid-v2 environment \citep{Todorov2012mujoco,brockman2016openai} with two major modifications. 
A state $s\in \mathbb{R}^{378}$ contains the positions of the different body parts as well as their velocities. In order to define an easily interpretable goal space corresponding to the Cartesian position of the torso, we also include the $x, y$ position of the torso in the states, while they are generally discarded. 
The point is that, in order to walk to infinity, the agent must learn to ignore these two inputs which makes this variant much more difficult than the original one. 
To compensate for this increased difficulty, the evaluation budget is reduced to $200$ control steps instead of $1000$. As a result, an optimal agent covers a forward distance of $\sim10$ meters instead of $\sim50$ meters. 
Demonstrations are obtained by an agent trained with the SAC algorithm on the original dense reward associated with the environment which is much easier than our sparse reward context. This original reward consists of three parts: a forward reward obtained by the agent for moving forward along the x-axis, a health reward received for each step spent by the agent with its torso above $1$ meter and a cost penalizing the controls and contacts. These demonstrations are divided into 15 goals by DCIL-II. Besides, the DCIL-II agent is only rewarded when the agent reaches desired goals corresponding to a desired $(x,y,z)$ position of the torso.

\subsubsection{Humanoid stand-up}

The \textit{Humanoid stand-up} environment is similar to the Humanoid locomotion environment. The only difference is that, instead of being initialized standing, the agent is laying down on its back at the beginning of an episode. 
Demonstrations are also obtained using SAC and the easier original dense reward function, which contains two parts: a height reward proportional to the position of the torso of the agent along the z-axis and a cost penalizing the controls. 
A demonstration corresponds to the Humanoid lifting its torso above $1.4$ meters. DCIL-II divides this demonstration into 12 goals. Similarly to the Humanoid locomotion task, DCIL-II only uses a sparse reward received when the agent reaches the desired $(x,y,z)$ position of the torso.  

\subsubsection{Cassie run}

The \textit{Cassie run} environment is a variant of the \textit{Humanoid locomotion} environment where the humanoid is replaced by a simulated Cassie robot which recently broke a running speed record\footnote{\url{https://today.oregonstate.edu/news/bipedal-robot-developed-oregon-state-achieves-guinness-world-record-100-meters}}.
In this environment, states $s\in \mathbb{R}^{67}$ contain the Cartesian positions and the orientations of each part of the bipedal platform (including the $x, y$ position of the robot as discussed in Section~\ref{sec:humanoid_locomotion}). 
A goal $g\in \mathbb{R}^{3}$ corresponds to the Cartesian position of the pelvis. DCIL-II only uses a sparse reward received by the agent when it reaches these low-dimensional goals. 
Demonstrations are obtained after $18M$ training steps using the TQC algorithm \cite{kuznetsov2020controlling} on a hand-designed dense reward function \cite{xpag}. 

\subsection{Baselines}

We compare DCIL-II to four baselines: 1) a random initialization of the weights of the policy, used as a lower bound on the performance of an agent in each environment;  2) a naive BC method using a single demonstration; 3) PWIL, a state-of-the-art IL algorithm in the Humanoid environment with a single demonstration; 4) DCIL-I.  

\subsection{Implementation details}
\label{sec:implementation_details}

The hyper-parameters for each environment are presented in Table~\ref{tab:hyper_parameters}. In the three environments, the observations are normalized with the mean and standard deviation of the observations gathered during an initialization phase where random actions are sampled. 
Moreover, during the SAC update, the target value is clipped in $[0, N_{goals}]$. This upper bound for the value corresponds to the maximum reward that an agent can obtain along a trajectory with a discount factor $\gamma = 1$. We found that clipping the value during the critic update greatly improved the stability of the agent and limited poor generalization of the critic network. The code of DCIL-II based on the XPAG library \cite{xpag} is provided at \url{https://github.com/AlexandreChenu/DCIL_XPAG}.

\begin{figure}[ht!]
     \centering
     \includegraphics[width=0.9\hsize]{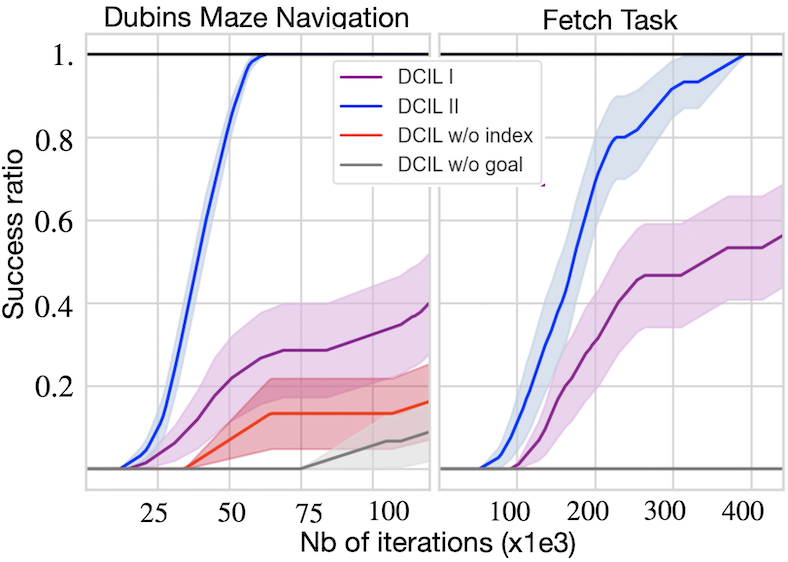}
    \caption{Comparing DCIL-II to three variants: we evaluate the success rates of DCIL-I, DCIL w/o index and DCIL w/o goal throughout training in the Dubins Maze and in the Fetch Task. The mean and standard deviation ranges over 15 seeds. The standard deviation is divided by four for more readability.}
    \label{fig:ablation_study}
\end{figure}

\subsection{Ablation study}
\label{sec:ablation_study}

Using the Dubins Maze and the Fetch environments, we compare DCIL-II to three variants inspired by the discussions in Sections~\ref{sec:DCIL_i_DCIL_ii}, \ref{sec:DCIL_ii_HRL} and \ref{sec:DCIL_i_RDSC}.
The first variant underlines the importance of combining the value propagation mechanism commonly used by DCIL-II and Option-Critic methods with GCRL and, in particular, a relabelling mechanism like HER (Section~\ref{sec:DCIL_ii_HRL}). 
In this variant called \textit{DCIL w/o goal}, the agent interacts directly with the MDP $\mathcal{M}_{seq}$ defined in Section~\ref{sec:problem_formulation}. There is no explicit goal-conditioning and relabelling mechanism. Therefore, the policy is only conditioned on the index of the goals which helps propagate the value associated with a goal into the value of the previous one. However, the lack of relabelling mechanism results in poor exploration.

Similarly to R-DSC (as discussed in Section~\ref{sec:DCIL_ii_HRL}), in the second variant called \textit{DCIL w/o index}, the agent interacts with a GC-MDP using a vanilla version of HER where success states are terminal. In this context, the agent benefits from efficient exploration. However, in the absence of reward propagation between successive goals, the agent is not encouraged to reach valid success states only, hampering sequential goal-reaching.

The third variant is DCIL-I. As explained in Section~\ref{sec:DCIL_i_DCIL_ii}, the main difference between DCIL-I and DCIL-II lies in the relabelling mechanism. In DCIL-I, relabelling is as in the original version of HER and success states are terminal. The agent only receives the reward bonus used to propagate the value from goal $g_{i}$ in $\tau_{\mathcal{G}}$ to the previous one $g_{i-1}$ when it reaches this precise goal $g_{i}$, not when it reaches an artificial goal created by HER. On the contrary, in DCIL-II, artificial successful transitions created by the relabelling mechanism also propagate the value from the next indices to the previous ones.

These differences in value propagation between DCIL w/o index, DCIL-I and DCIL-II are illustrated in \figurename~\ref{fig:comp_value_propagation}. DCIL-II is able to increase the value of valid success states for any intermediate goals (original goal from $\tau_{\mathcal{G}}$ and relabeled goals). As a result, an agent trained with DCIL-II learns how to navigate the entire \textit{Dubins Maze} and to solve the Fetch task faster than with any of the variants (see \figurename~\ref{fig:ablation_study}).


\begin{figure}[ht!]
     \centering
     \includegraphics[width=\hsize]{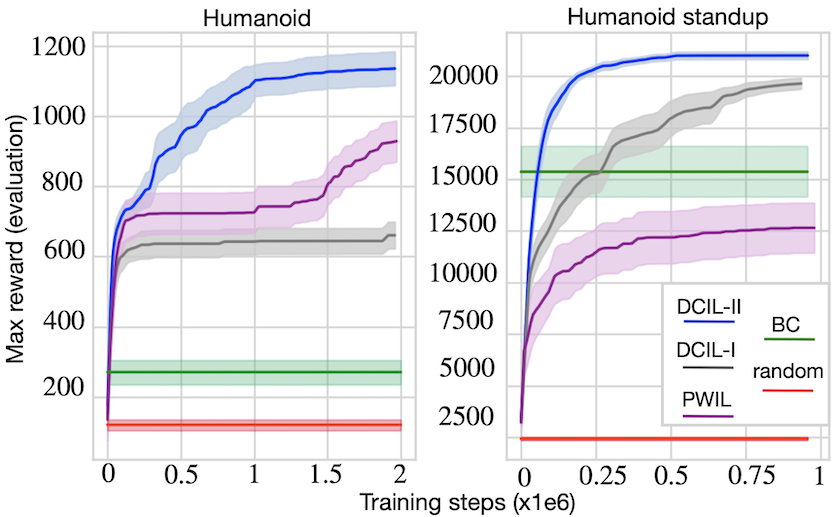}
    \caption{Comparing DCIL-II to a random policy, BC, PWIL and DCIL-I. We evaluate the maximum return obtained by the agent throughout training in the Humanoid locomotion and Humanoid stand-up environments. The mean and standard deviation ranges over 15 seeds. The standard deviation is divided by two for more readability. }
    \label{fig:results_dcil_ii} 
\end{figure}

\subsection{Comparison to baselines}

In this section, we compare DCIL-II to all baselines in the two humanoid environments. 

\begin{figure*}[t!]
     \centering
     \includegraphics[width=\hsize]{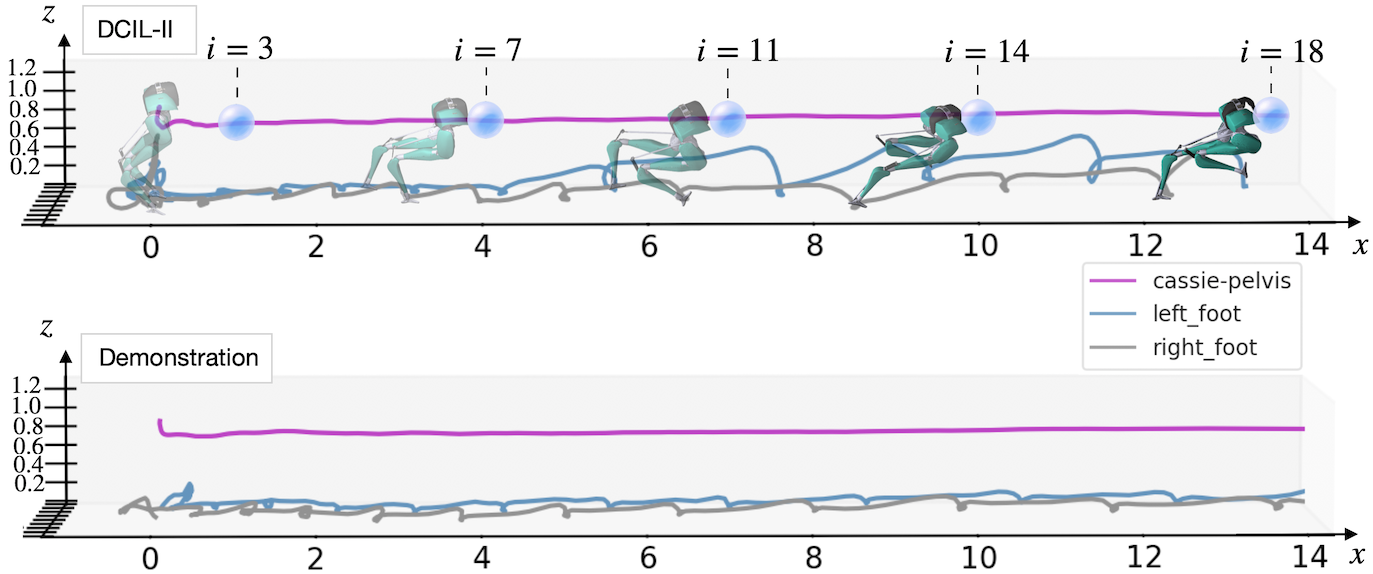}
    \caption{Visualization of sequential goal reaching in the \textit{Cassie run} environment. In this environment, a sequence of 19 goals $\tau_{\mathcal{G}}$ corresponding to successive Cartesian positions of the pelvis is used by DCIL-II to learn the behavior using only 707K training steps on average. Only five goals are shown here. One can see that, though the straight pelvis trajectory is well reproduced, the obtained feet trajectories differ widely from the demonstrated ones.}
    \label{fig:cassie_run_visu}
\end{figure*}

\figurename~\ref{fig:results_dcil_ii} presents the average maximum return gathered by an agent over evaluation trajectories in both environments. 
During the evaluation trajectory, the agent is reset in the first state of the demonstration. At each control step, actions are selected according to the mean of the action distribution returned by the policy network. 
The return is computed according to the original reward for each environment. However, none of the methods trained directly using this dense reward.

In the locomotion task, DCIL-II is the only method able to reach an average maximum return superior to 1100 within 2M training steps. This reward corresponds to an agent walking a distance of $\sim 10m$.
As discussed in the presentation of the environment, the added $x, y$ positions of the agent forces non goal-conditioned agents (as in PWIL) to ignore these two inputs. This results in a slower increase of the return of PWIL compared to the results presented in the original paper \cite{dadashi2020primal}.  

\begin{figure}[ht!]
     \centering
     \includegraphics[width=\hsize]{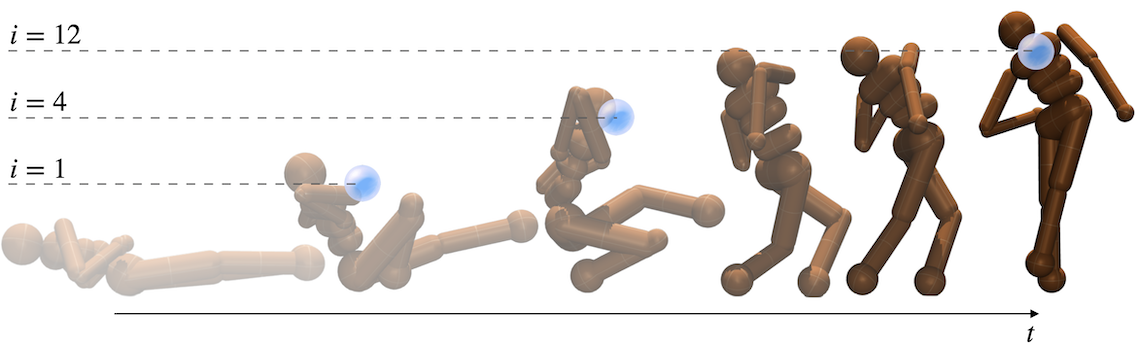}
    \caption{Visualization of sequential goal reaching in the \textit{Humanoid stand-up} environment. In this environment, a sequence of 12 goals $\tau_{\mathcal{G}}$ corresponding to successive Cartesian positions of the torso is used by DCIL-II to learn the behavior using only 253K training steps. Only three goals are shown here.}
    \label{fig:humanoid_standup_visu}
\end{figure}

In the stand-up task, both DCIL-I and DCIL-II manage to achieve large returns around 20000 which correspond to an agent standing up (see \figurename~\ref{fig:humanoid_standup_visu}). Nevertheless, DCIL-II achieves such returns much faster than DCIL-I.

BC performs surprisingly well in this task. This is mainly due to the fact that the entire trajectory of the agent is mostly conditioned by the very first actions. Thus, the often problematic distribution shift faced by BC \cite{pmlr-v9-ross10a} has less impact in this environment. 

\subsection{Using DCIL-II to control a simulated Cassie bipedal platform}

In this environment, we want to learn a controller to make Cassie walk straight. DCIL-II manages to learn a walking behavior using only a sequence of low-dimensional goals corresponding to a succession of Cartesian positions of the pelvis.  
Only the position of this part of the robot is constrained during locomotion, the rest of the robot is free to follow trajectories that widely differ from the demonstration. 
\figurename~\ref{fig:cassie_run_visu} illustrates this property. Indeed, trajectories of the feet deviate a lot  from the demonstration. 
In addition, at the beginning of the trajectory, after being initialized slightly above the ground in a straight position, the simulated robot learns to fall on its feet by folding and unfolding its legs. These walking and stabilisation behaviours are totally different from and more complex than the demonstrated behaviours. Nevertheless, they still satisfy the desired successive positions of the pelvis.
Thus, DCIL-II learns a control policy to reach 19 successive goals using $707K \pm 144K$ training steps. By reaching these goals sequentially, the system covers a forward distance of 15 meters with a stable z-position of the pelvis.


\section*{Discussion \& Conclusion}
\label{sec:discussion}
\begin{table*}[ht!]
\centering
\begin{tabular}{ |M{3cm}|M{2cm}M{2cm}M{2cm}M{2cm}M{2cm}|  }
 \hline
 Hyper-parameters & \textit{Dubins Maze} & \textit{Fetch} & \textit{Humanoid locomotion} & \textit{Humanoid stand-up} & \textit{Cassie run}\\
 \hline
 \rowcolor{DarkGray}
 Critic hidden size  & $[400,300]$ & $[512,512,512]$ & $[512,512,512]$ & $[512,512,512]$ & $[512,512,512]$   
 \\
 \rowcolor{DarkGray}
 Policy hidden size & $[400,300]$ & $[512,512,512]$ & $[512,512,512]$ & $[512,512,512]$ & $[512,512,512]$   
 \\
 \rowcolor{DarkGray}
 Activation functions & ReLU & ReLU & ReLU & ReLU & ReLU
 \\
 \rowcolor{DarkGray}
 Batch size & 256 & 256 & 64 & 64 & 64
 \\
 \rowcolor{DarkGray}
 Discount factor & 0.9 & 0.98 & 0.98 & 0.98 & 0.98
 \\
 \rowcolor{DarkGray}
 Entropy coefficient & $1\times10^{-3}$ & $1\times10^{-4}$ & $1\times10^{-4}$ & $1\times10^{-4}$ & $1\times10^{-4}$ 
 \\
 \rowcolor{DarkGray}
 Critic lr & $1\times10^{-3}$ & $1\times10^{-4}$ & $3\times10^{-4}$ & $3\times10^{-4}$ & $3\times10^{-4}$
 \\
 \rowcolor{DarkGray}
 Policy lr & $1\times10^{-3}$ & $1\times10^{-4}$ & $3\times10^{-4}$ & $3\times10^{-4}$ & $3\times10^{-4}$
 \\
 $\epsilon_{success}$ & 0.1 & 0.05 & 0.05 & 0.05 & 0.05 
 \\
 $\epsilon_{dist}$ & 1 & 0.5 & 0.5 & 0.3 & 0.75
 \\
 $\tau_{\mathcal{G}}$ size & 22 & 12 & 15 & 12 & 19
 \\
 Training rollout budget & 25 &  100 & 100 & 100 & 100
 \\
 \hline
\end{tabular}
\caption{Hyper-parameters used for SAC (grey) and DCIL-II (white). The Hyper-parameters relative to SAC for \textit{Humanoid locomotion} are extracted from \cite{rl-zoo} and reused for \textit{Humanoid stand-up} and \textit{Cassie run}.}
\label{tab:hyper_parameters}
\end{table*}

In this paper, we have tackled the problem of learning complex robotics behavior using a single demonstration. 
We leveraged a sequential inductive bias, resulting in the design of an augmented GCRL framework where the agent must successively reach low-dimensional goals in order to achieve a complex task. In this framework, the agent is encouraged to achieve each goal via states that are compatible with reaching the subsequent goals. 
Based on this framework, we presented the DCIL-II algorithm and we have shown that it can learn control policies significantly faster than state-of-the-art imitation learning methods when addressing complex articulated behaviors with simulated and underactuated robots. 

Similarly to DCIL-I, one of the main limitations of DCIL-II under the perspective of applying it to a real robot is the assumption that the agent can be reset in some states selected in the demonstrated trajectory. Rather than being reset, we would like the robot to come back to a required state using its own control law, as in \cite{ecoffet2021first, gupta2021reset}. This will require enriching our framework with the capability to learn a large set of unrelated skills, which is left for future work.


\section*{Acknowledgements}
This work was partially supported by the French National Research Agency (ANR), Project ANR-18-CE33-0005 HUSKI and was performed using HPC resources from GENCI-IDRIS (Grant 2022-A0111013011).
This work also received funding from the European Commission’s Horizon Europe Framework Programme under grant agreement No 101070381 (PILLAR-robots project).



%
\bibliographystyle{IEEEtran}

%





\end{document}